\documentclass[sigconf]{acmart}
\usepackage{ bbold }
\usepackage{ stmaryrd }
\usepackage{amsmath}
\usepackage{multirow}
\usepackage{enumitem}

\AtBeginDocument{%
  }

\setcopyright{acmlicensed}
\copyrightyear{2018}
\acmYear{2018}
\acmDOI{XXXXXXX.XXXXXXX}
\acmConference[Conference acronym 'XX]{Make sure to enter the correct
  conference title from your rights confirmation email}{June 03--05,
  2018}{Woodstock, NY}
\acmISBN{978-1-4503-XXXX-X/2018/06}

\makeatletter
\DeclareRobustCommand\onedot{\futurelet\@let@token\@onedot}
\def\@onedot{\ifx\@let@token.\else.\null\fi\xspace}

\makeatother

\begin{document}

\title{Traj-Transformer: Diffusion Models with Transformer for GPS Trajectory Generation}

\author{Zhiyang Zhang}
\email{zzhang18@wpi.edu}
\affiliation{%
  \institution{Worcester Polytechnic Institute}
   \country{USA}
}

\author{Ningcong Chen}
\email{nchen3@wpi.edu}
\affiliation{%
  \institution{Worcester Polytechnic Institute}
   \country{USA}
}

\author{Xin Zhang}
\email{xzhang19@sdsu.edu}
\affiliation{%
  \institution{San Diego State University}
  \country{USA}
  }

\author{Yanhua Li}
\email{yli15@wpi.edu}
\affiliation{%
  \institution{Worcester Polytechnic Institute}
   \country{USA}
}

\author{Shen Su}
\email{sushen@gzhu.edu.cn}
\affiliation{%
  \institution{Guangzhou University}
   \country{China}
}

\author{Hui Lu}
\email{luhui@gzhu.edu.cn}
\affiliation{%
  \institution{Guangzhou University}
   \country{China}
}

\author{Jun Luo}
\email{jluo1@lenovo.com}
\affiliation{%
 \institution{Lenovo Group Limited}
 \country{Hong Kong}
 }


\renewcommand{\shortauthors}{Anonymous}

\begin{abstract}
The widespread use of GPS devices has driven advances in spatiotemporal data mining, enabling machine learning models to simulate human decision-making and generate realistic trajectories—addressing both data collection costs and privacy concerns.

Recent studies have shown the promise of diffusion models for high-quality trajectory generation. However, most existing methods rely on convolution based architectures (e.g. UNet) to predict noise during the diffusion process, which often results in notable deviations and the loss of fine-grained street-level details due to limited model capacity.
In this paper, we propose Trajectory Transformer (Traj-Transformer), a novel model that employs a transformer backbone for both conditional information embedding and noise prediction. We explore two GPS coordinate embedding strategies—location embedding and longitude-latitude embedding—and analyze model performance at different scales.

Experiments on two real-world datasets demonstrate that Traj-Transformer significantly enhances generation quality and effectively alleviates the deviation issues observed in prior approaches.
\end{abstract}




\received{20 February 2007}
\received[revised]{12 March 2009}
\received[accepted]{5 June 2009}

\maketitle

\begin{figure}
   \includegraphics[scale=0.19]{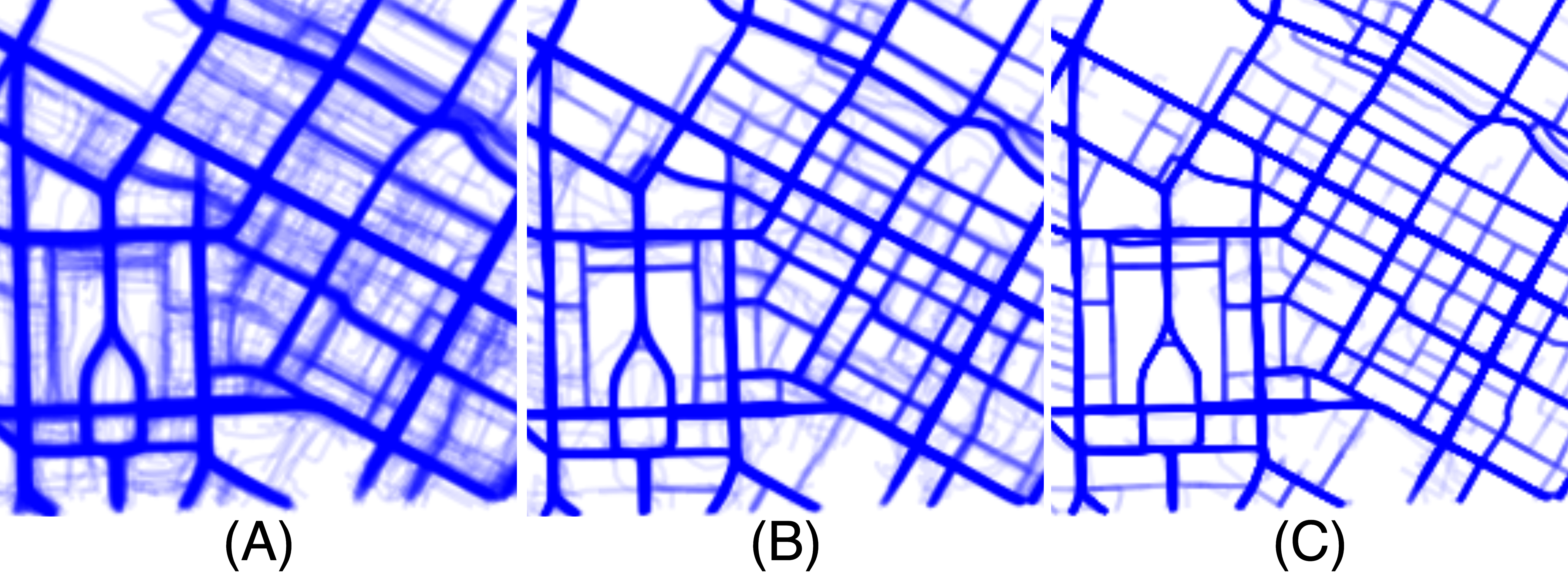}
   \centering
   \caption{Trajectories generated in a high density urban region, using models that are directly trained on raw GPS trajectories without road network. (A): Convolution-based models struggle to reconstruct the street structure. (B): Our model preserves fine-grained, street-level details, leading to significantly improved generation quality. (C): Raw GPS trajectories collected by GPS devices.}
   \label{teaser}
\end{figure}

\section{Introduction}
GPS trajectory data contains vast valuable information that can be effectively utilized across a wide range of applications, such as urban planning \cite{ruan2020learning},
intelligent transportation systems \cite{ma2013t}, human mobility analysis \cite{liang2021modeling, wang2021trajectory}, and public safety monitoring \cite{gao2017identifying}.
Despite its potential, raw GPS trajectory data poses challenges, including privacy risks and the high cost of large-scale data collection. Thus, generating human-like trajectories to replace real data is crucial for real-world applications.

Generating GPS trajectories presents several practical challenges. First, when dealing with a large volume of trajectories, the distribution becomes highly complex due to variations in population density and the intensity of human activity across different urban regions \cite{zhao2022multi, zhu2022cross}. This complexity demands a robust modeling approach capable of capturing such diverse patterns. Second, on an individual level, each trajectory exhibits unique characteristics, stemming from the inherently stochastic nature of human behavior, which makes accurate prediction particularly difficult \cite{zheng2015trajectory, liang2021modeling}. Diffusion models \cite{sohl2015deep, ho2020denoising, song2020score}, a promising approach in generative AI field, exhibit impressive capability to capture the complicated data distribution.

Zhu et al. \cite{zhu2023difftraj} were the first to explore trajectory generation using diffusion models. In their work, GPS trajectories are represented as two-dimensional tensors. Following the standard procedure in diffusion models, Gaussian noise is added onto the trajectory tensor to perturb and a model called Traj-UNet was proposed to predict the noise during the diffusion sampling process. In a subsequent study \cite{zhu2024controltraj}, they introduce GeoUNet, a model that incorporates road embeddings to enhance generation quality. In order to get the road embedding, an auto-encoder model need to be pretrained. Although incorporating road information can improve the results, it incurs a two-stage training process: one for learning road embeddings and another for noise prediction. Another similar approach \cite{wei2024diff} employs the Node2Vec algorithm \cite{grover2016node2vec} to obtain road embeddings.

All of the aforementioned methods are based on standard UNet architectures \cite{ronneberger2015u} with minor variations such as the use of ResNet blocks \cite{he2016deep}, self-attention \cite{vaswani2017attention} mechanisms in intermediate layers, and dilated convolutions \cite{yu2015multi, zhang2021towards} to capture patterns within multiple ranges. In \cite{zhu2023difftraj}, the authors adopt a CNN-based architecture for computational efficiency. However, we observe that UNet-based models significantly limit the quality of the generated trajectories. For example, a substantial portion of the generated trajectories deviate from the road network—an issue not present in the training data—and fine-grained details are often lost in densely populated urban regions. We found the UNet inductive bias help to learn the trajectory distribution but not sufficient to learn better. To improve the generation quality, we turn to transformers \cite{vaswani2017attention}, which hold more relaxed inductive bias than convolutional architectures \cite{lavie2024towards}. Recent research has demonstrated their strong capability in processing continuous data \cite{dosovitskiy2020image}, while transformer was originally proposed for handling discrete data such as nature language processing \cite{vaswani2017attention}.

In this paper, we propose a transformer-based model for GPS trajectory generation (Traj-Transformer). In terms of model design, we investigate two different strategies for GPS point embedding. With even one-quarter parameters, our model can significantly improve the generation quality, preserve the fine-grained street-level details and effectively mitigate the deviation issues observed in previous approaches (\textbf{Figure} \ref{teaser}), in which the models use convolutional based UNet as their backbone.

To summarize, the main contributions of this work are as follows:
\begin{itemize}[nosep, leftmargin=*]
    \item We propose Traj-Transformer, a Transformer-based model for noise prediction in GPS trajectory generation, and demonstrate that it significantly improves generation quality.
    \item We investigate two GPS point embedding strategies and show that embedding longitude and latitude separately leads to better performance in GPS trajectory generation task.
    \item We validate our approach on two real-world datasets, demonstrating the capability of transformer-based models to generate high-quality trajectories, supported by both quantitative metrics and qualitative visualizations.
\end{itemize}

\section{Preliminary}
In this section, we formally define the GPS trajectory generation problem, introduce the notations used throughout the paper, and provide a brief overview of the diffusion model framework.
\subsection{Definitions}

\noindent\textbf{Definition 1 (GPS Trajectory).} A N-length GPS trajectory $x = \{p_{1}, p_{2}, \cdots ,p_{N}\}$ is a sequence of GPS points, where $p_i = [\text{lon}_i, \text{lat}_i]$ represents the longitude and latitude. The entire trajectory $x$ records the movement of an object over time.

\noindent\textbf{Problem Definition. } Given a set of real-world GPS trajectories $\mathcal{D} = \{x_{1}, x_{2}, \cdots, x_{n}\}$, where $x_{i} = \{p_{1}^{i}, p_{2}^{i}, \cdots ,p_{N}^{i}\}$ is the $i$-th trajectory. The objective of GPS trajectory generation is to learn a generative model $G_{\theta}$ with parameters $\theta$ that can approximate the distribution of real-world trajectories such that the generated trajectories $\mathcal{G}$ from $G_{\theta}$ preserve key spatial-temporal characteristics, distribution of real trajectories and movement diversity, i.e.
\begin{align}
    \max_\theta \sum_{x_{i}\in\mathcal{D}} \log P_\theta(x_i), \text{s.t. } \tilde{x}\in \mathcal{G},  \forall \tilde{x}\sim P_\theta(x). 
\end{align}
In practice, the generated trajectories $\mathcal{G}$ should remain useful for downstream applications and analysis, like traffic flow prediction \cite{maerivoet2005traffic}, urban planning \cite{ruan2020learning}, and human mobility analysis \cite{gao2017identifying}.

\subsection{Denoising Diffusion Probability Models}
Denoising Diffusion Probability Models (DDPMs) \cite{ho2020denoising}, a family of generative models, first proposed by Sohl-Dickstein et al. \cite{sohl2015deep} have demonstrated remarkable performance in producing high-quality data \cite{dhariwal2021diffusion}. The essence of DDPMs is perturbing clean data with Gaussian noise so that the data distribution becomes normal distribution, and we learn a reverse process that can recover the data distribution from tractable normal distribution which is easy to sample. This is known as forward and reverse processes.

\noindent\textbf{Forward process.} The forward process is a Markov chain \cite{ho2020denoising}. We perturb clean data step by step with Gaussian noise added on previous step. The maximum perturbation step is $T$. Formally, Let $x_{0}\sim P_{data}$, the forward process at timestep $t$ can be defined as:
\begin{align}
x_{t} = \sqrt{1-\beta_{t}} x_{t-1} + \sqrt{\beta_{t}} \epsilon \sim \mathcal{N}(x_{t};\sqrt{1-\beta_{t}} x_{t-1}, \beta_{t}\boldsymbol{I}), \label{eq:forward0}
\end{align}
where $\beta_{t} \in (0, 1)$ is the noise schedule and $\epsilon \sim \mathcal{N}(0,\boldsymbol{I})$.

\noindent For efficient computation, (\ref{eq:forward0}) can be reparameterized into:
\begin{align}
x_{t} = \sqrt{\bar{\alpha}_{t}} x_{0} + \sqrt{1 - \bar{\alpha}_{t}} \epsilon \sim \mathcal{N}(x_{t};\sqrt{\bar{\alpha}_{t}} x_{0}, (1 - \bar{\alpha}_{t})\boldsymbol{I}), \label{eq:forward1}
\end{align}
where $\bar{\alpha}_{t}=\prod_{i=1}^{t}(1-\beta_{i})$

\noindent\textbf{Reverse process.} Reverse process is to generate clean data with a denoising process starting from noise $x_{T} \sim \mathcal{N}(0,\boldsymbol{I})$. Formally, denoising process recovers $x_{t-1}$ from $x_{t}$ with probability:
\begin{align}
q(x_{t-1}|x_{t}) \sim \mathcal{N}(x_{t-1}; \mu_{t}(x_{t}), \sigma_{t}^2\boldsymbol{I}), \label{eq:backward0}
\end{align}
where $\mu_{t}(x_{t})=\frac{1}{\sqrt{\alpha_{t}}}(x_{t}-\frac{1-\alpha_{t}}{\sqrt{1-\bar{\alpha}_{t}}}\epsilon)$, $\sigma_{t}^2=\frac{(1-\bar{\alpha}_{t-1})(1-\alpha_{t})}{1-\bar{\alpha}_{t}}$.

\noindent Here, in practice, the noise $\epsilon$ is unknown and will be estimated by a neural network $\epsilon_{\theta}$ with $x_{t}$ and timestep $t$ as input.

\noindent The full diffusion process typically involves thousands of timesteps \cite{ho2020denoising}, so that the step-by-step denoising procedure can be extremely slow during sampling. To accelerate the denoising process, Denoising Diffusion Implicit Models (DDIMs) \cite{song2020denoising} is proposed. DDIMs are a class of non-Markovian diffusion processes that share the same training objective as standard Denoising Diffusion Probabilistic Models (DDPMs), but allow for a more efficient and flexible sampling process. In the reverse process of DDIMs, the denoise sample $x_{t-1}$ is computed from $x_{t}$ as follows:
\begin{align}
x_{t-1} = \sqrt{\alpha_{t-1}} &\left( {\frac{x_t - \sqrt{1-\alpha_t} \epsilon_\theta(x_t, t)}{\sqrt{\alpha_t}}} \right) + \nonumber \\
&{\sqrt{1-\alpha_{t-1} - \sigma_t^2} \cdot \epsilon_\theta(x_t, t)} + {\sigma_t \epsilon_t}. \label{eq:backward1}
\end{align}
When $\sigma_t = \sqrt{\frac{1 - \alpha_{t-1}}{1 - \alpha_t}} \sqrt{1 - \frac{\alpha_t}{\alpha_{t-1}}}$, DDIMs denoising (\ref{eq:backward1}) becomes DDPMs denoise (\ref{eq:backward0}).

\noindent In (\ref{eq:backward1}), denoising for one step is not necessary, one can sample every $\lceil \frac{T}{S} \rceil$ steps $(S < T)$, effectively reducing the total number of denoising steps from $T$ to $S$.

\begin{figure}
   \includegraphics[scale=0.75]{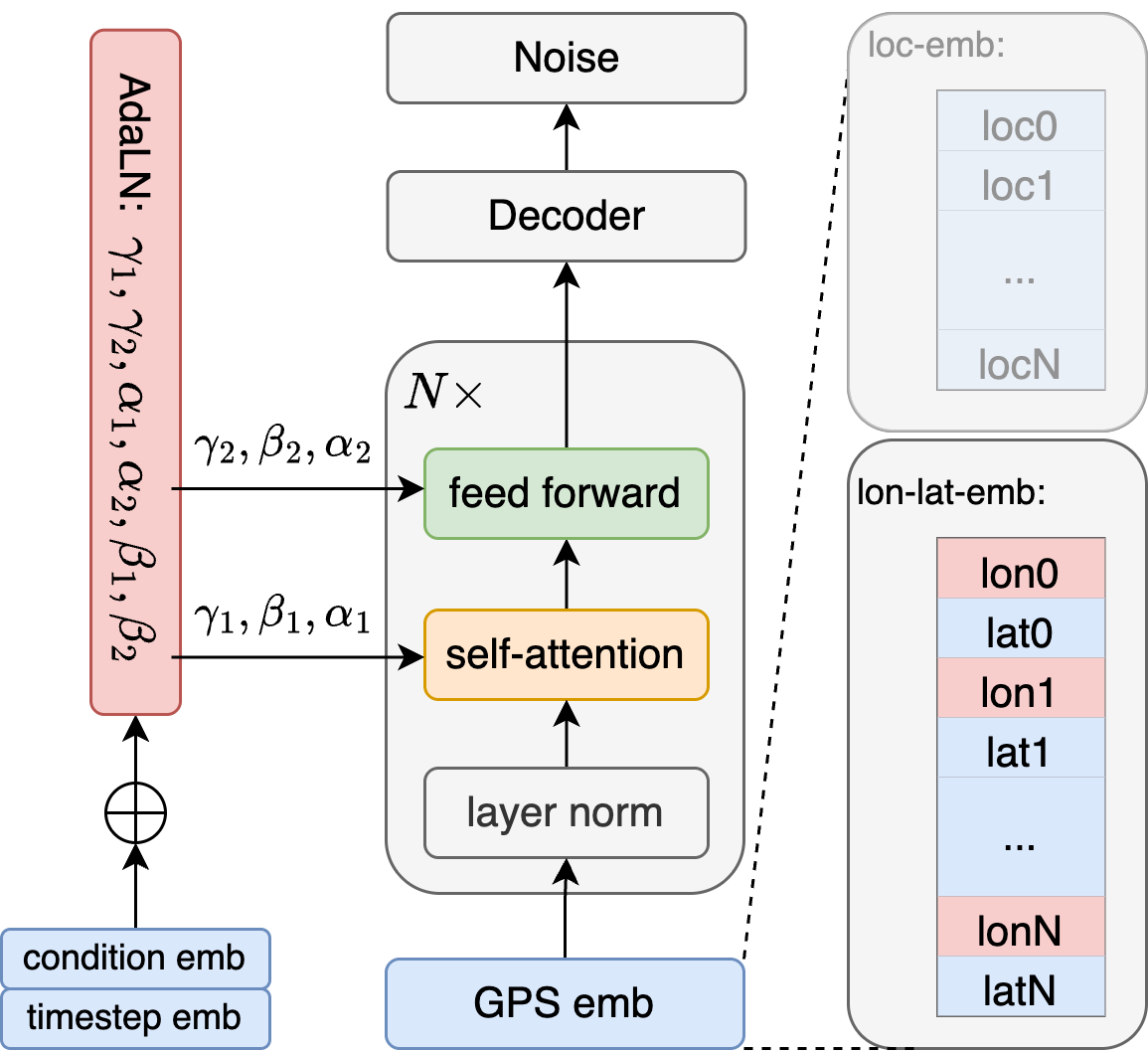}
   \centering
   \caption{Architecture of the Trajectory Transformer (Traj-Transformer).
The model takes GPS trajectories as input and supports two alternative embedding strategies for GPS points: (1) loc-emb, which computes an embedding for each location, and (2) lon-lat-emb, which independently embeds longitude and latitude coordinates. These embeddings are then fed into a Transformer backbone, which serves as the core of our model. To enable conditional generation, both the generation conditions and diffusion timesteps are injected into the transformer layers using an adaptive layer norm (adaLN). After passing through the decoder, the model produces noise predictions that are used in the diffusion reverse process to denoise.}
   \label{model_arch}
\end{figure}

\section{Trajectory Transformer}
In this section, we introduce our model Trajectory Transformer (Traj-Transformer), that leverages a Transformer backbone for GPS trajectory generation. We describe the model from three key perspectives: (1) the embedding strategies for GPS points, (2) the architecture for noise prediction and conditional information injection, and (3) the model configurations used in our experiments.
\subsection{GPS point embedding}
As introduced in \textbf{definition 1}, for a N-length GPS trajectory $x = \{p_{1}, p_{2}, \cdots ,p_{N}\}$ and $p_i = [\text{lon}_i, \text{lat}_i]$, we can follow the embedding in \cite{zhu2023difftraj}, treating longitude and latitude together so as a N-length GPS trajectory will be embedded into $X^{N\times D}$ ($D$ is the embedding dimension), in which
\begin{align}
X_{i} = emb([\text{lon}_i, \text{lat}_i]).
\end{align}
Here, Longitude and latitude together represents a location on map and finally embedded into one single vector. We refer this embedding as location embedding or loc-emb for short (\textbf{Figure} \ref{model_arch}).

An alternative way to embed the GPS point is to treat longitude and latitude separately. Longitude and latitude follow different embedding layers and finally a N-length trajectory will be embedded into $X^{2\times N\times D}$ ($D$ is the embedding dimension), in which
\begin{align}
X_{2i} = emb_{lon}(\text{lon}_i), \\
X_{2i+1} = emb_{lat}(\text{lat}_i),
\end{align}
Where $emb_{lon}$ and $emb_{lat}$ represent independent embedding layer for longitude and latitude respectively. As the embedding process suggests, we refer this embedding as longitude latitude embedding or lon-lat-emb for short (\textbf{Figure} \ref{model_arch}).

In loc-emb, 1D-positional encoding ($d$ is the embedding dimension) \cite{vaswani2017attention}:
\begin{align}
PE(n, d) = [sin(10^{\frac{0*4}{d/2}}*n), \dots, sin(10^{\frac{(d/2-1)*4}{d/2}}*n), \nonumber \\
cos(10^{\frac{0*4}{d/2}}*n), \dots, cos(10^{\frac{(d/2-1)*4}{d/2}}*n)]
\end{align}
is sufficient, but in lon-lat-emb, the consecutive two embeddings are longitude and latitude from the same location. We use 2D-positional encoding to help model distinguish longitude, latitude and location. Formally, the 2D-positional encoding for the $n$-th GPS point is:
\begin{align}
PE_{lon}(n, d) = [PE(0, d/2), PE(n, d/2)], \\
PE_{lat}(n, d) = [PE(1, d/2), PE(n, d/2)].
\end{align}
Here, the fixed identifiers (0 and 1) distinguish longitude and latitude, while the second half of the vector encodes temporal position within the sequence.

When using the lon-lat-emb strategy, longitude and latitude are embedded separately, allowing their representations to interact through dot-product operations in the self-attention operation. Intuitively, this embedding scheme preserves more spatial information, which is beneficial for capturing fine-grained, street-level details in trajectory data. We compare the effectiveness of loc-emb and lon-lat-emb in the experimental section to evaluate their impact on generation performance.

\begin{table}
       \centering
       \caption{Details of model configs. \textnormal{\textbf{T}iny, \textbf{S}mall, \textbf{B}ase, and \textbf{L}arge correspond to one-quarter, one-half, equal, and double to the UNet based model size.} }
       \renewcommand{\arraystretch}{1.2} 
       \begin{tabular}{ c c c c c } \hline
       Model & Layer $n$ & Dim $d$ & Head $h$ & Size\\ \hline
              T & 6 & 192 & 6 & 4.5M\\
              S & 12 & 192 & 6 & 8.5M\\
              B & 6 & 384 & 6 & 17M\\
              L & 12 & 384 & 6 & 33M\\ \hline
              \end{tabular}
        \vspace{0.5em}
        \label{tab:model_configs}
\end{table}

\begin{table*}
   \centering
   \caption{Performance comparison of different model sizes \& embeddings (model symbols correspond to Table \ref{tab:model_configs}).}
   \renewcommand{\arraystretch}{1.2} 
   \begin{tabular}{l c c c c c c | c c c c } \hline
      \multirow{2}{*}{Methods} & \multirow{2}{*}{Size} & \multirow{2}{*}{Gflops} & \multicolumn{4}{c|}{Chengdu}& \multicolumn{4}{c}{Xi'an}\\ \cline{4-11}
                & & & Density ($\downarrow$) & Trip ($\downarrow$) & Length ($\downarrow$) & Pattern ($\uparrow$) &
                Density ($\downarrow$) & Trip ($\downarrow$) & Length ($\downarrow$) & Pattern ($\uparrow$) \\ \hline
         T/loc & 4.5M & 0.5 & 0.0649 & 0.3554 & 0.0195 & 0.732 & 0.0492 & 0.3421 & 0.0223 & 0.764 \\
         S/loc & 8.5M & 1.1 & 0.0538 & 0.3192 & 0.0175 & 0.758 & 0.0423 & 0.3152 & 0.0220 & 0.770 \\
         B/loc & 17M & 2.1 & 0.0454 & 0.2985 & 0.0152 & 0.775 & 0.0371 & 0.2734 & 0.0212 & 0.790 \\
         L/loc & 33M & 4.3 & 0.0352 & 0.2742 & 0.0125 & 0.813 & 0.0332 & 0.2221 & 0.0198 & 0.811 \\ \hline
         T/lon-lat & 4.5M & 1.1 & 0.0568 & 0.3318 & 0.0189 & 0.760 & 0.0457 & 0.3221 & 0.0220 & 0.775 \\
         S/lon-lat & 8.5M & 2.1 & 0.0473 & 0.3040 & 0.0168 & 0.770 & 0.0395 & 0.2956 & 0.0219 & 0.788 \\
         B/lon-lat & 17M & 4.3 & 0.0376 & 0.2882 & 0.0143 & 0.812 & 0.0352 & 0.2431 & 0.0207 & 0.793 \\
         L/lon-lat & 33M & 8.5 & \textbf{0.0303} & \textbf{0.2688} & \textbf{0.0118} & \textbf{0.828} & \textbf{0.0295} & \textbf{0.1953} & \textbf{0.0189} & \textbf{0.830} \\ \hline
      \end{tabular}
   \label{Tab:results1}
\end{table*}

\begin{figure*}
   \includegraphics[scale=0.27]{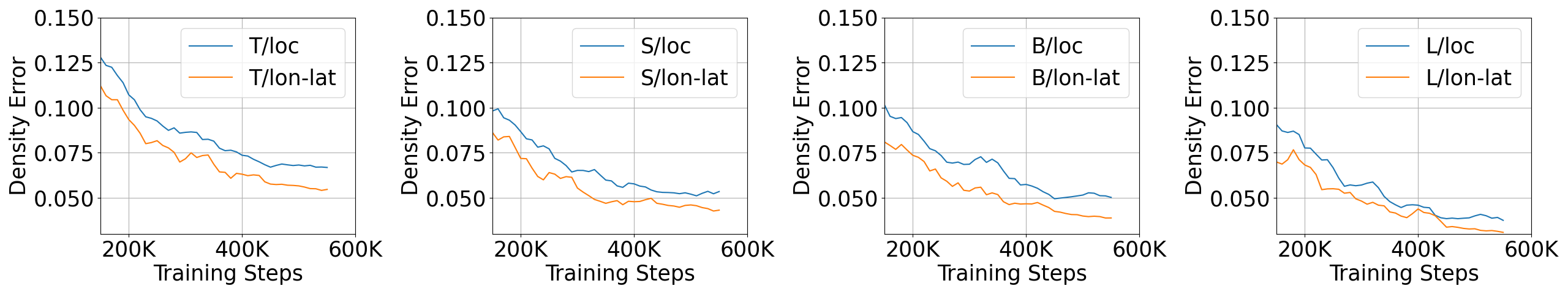}
   \centering
   \caption{Density error over training steps on Chengdu. Models using lon-lat embeddings consistently outperform those with loc embeddings throughout all training stages. Similar trends are observed when training on Xi’an.}
   \label{embedding_training_curve}
\end{figure*}

\subsection{Condition, Timestep \& Decoder}
When predicting noise with neural network, it's essential to input $x_t$ along with the corresponding timestep $t$, which allows the model to perceive the diffusion step. Additionally, conditional information $c$ is supplied to guide the generation toward the desired target. Finally, the model output the noise estimation $\epsilon_\theta(x_t, t, c)$.

Noisy data $x_t$, timestep $t$ and condition $c$ represent distinct modalities, making their integration non-trivial. While cross-attention mechanisms can be used to merge these inputs, they are often computationally inefficient (quadratic complexity). Instead, we adopt the adaptive layer norm (adaLN) approach, as proposed in \cite{perez2017visual} and further demonstrated in diffusion models \cite{peebles2023scalable}, which provides an effective and efficient (linear complexity) way to incorporate all three components into the model (\textbf{Figure} \ref{model_arch}).

In adaLN \cite{perez2017visual, peebles2023scalable}, conditional information and diffusion timesteps are injected into the model both before and after the self-attention and feedforward layers (\textbf{Figure} \ref{model_arch}). This is achieved through learned scaling and shifting operations, where the inputs are modulated by scale parameters ($\gamma, \alpha$) and shift parameters($\beta$), respectively:
\begin{align}
\alpha_1\odot SelfAttention((1 + \gamma_1)\odot LayerNorm(x) + \beta_1), \\
\alpha_2\odot FeedForward((1 + \gamma_2)\odot LayerNorm(x) + \beta_2),
\end{align}
Where $\gamma_i, \alpha_i$, and $\beta_i$ are calculated on the sum of the embedding vectors of $t$ and $c$ with linear layers $W_i^{\gamma}, W_i^{\alpha}$, and $W_i^{\beta}$:
\begin{align}
y &= t + c, \\
\gamma_i &= W_i^{\gamma}y, \alpha_i = W_i^{\alpha}y, \beta_i = W_i^{\beta}y.
\end{align}

When initializing, setting scale($\gamma, \alpha$) and shift($\beta$) to zero, which lead the initial neural neural network to an identity map, was proved to be an efficient strategy for training in practice \cite{goyal2017accurate}.

After the final layer, the output is decoded into a noise prediction matching the shape of the input. The decoder employs distinct strategies for loc-emb and lon-lat-emb: it linearly projects each token into 2 dimensions for loc-emb, and into 1 dimension for lon-lat-emb, respectively. The outputs are then rearranged to restore the original input shape, producing the final noise prediction.

\subsection{Model Size}
Based on the typical parameter count of UNet-based models for GPS trajectory generation (approximately 16 million), we design four model configurations by scaling the number of layers and embedding dimensions. These configurations correspond to one-quarter (\textbf{T}iny), one-half (\textbf{S}mall), equal (\textbf{B}ase), and double (\textbf{L}arge) the size of the UNet baseline. Each configuration supports both loc-emb and lon-lat-emb embedding strategies. Details of these configurations are summarized in \textbf{Table \ref{tab:model_configs} }.

\begin{figure*}
   \includegraphics[scale=0.36]{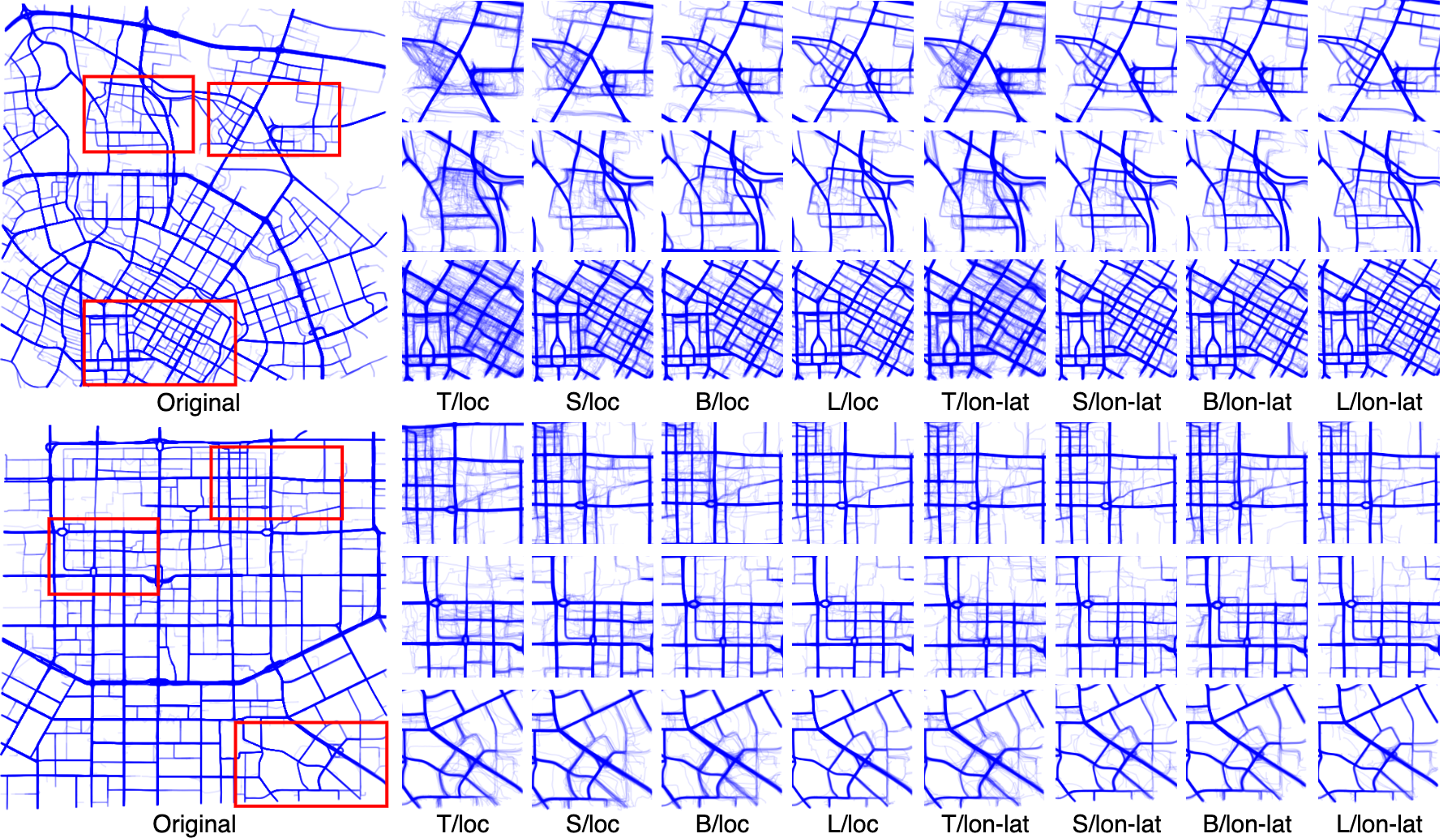}
   \centering
   \caption{Visualizations of two cities (top: Chengdu; bottom: Xi'an) from different models. Red rectangles indicate high-density urban regions where performance differences among models are most pronounced. To facilitate direct comparison, each highlighted patch presents a magnified view of the corresponding region generated by different models. Lower-performing models tend to overlook fine-grained, street-level structures, whereas higher-performing models more accurately capture and preserve these intricate details. All visualizations depict 5,000 trajectories, rendered without any visual post-processing.}
   \label{compare1}
\end{figure*}

\section{Experimental Setup}
\noindent\textbf{Datasets.} To comprehensively evaluate the performance of our proposed model, we conduct experiments on two large-scale GPS trajectory datasets\footnote{\url{https://outreach.didichuxing.com/}}, each capturing vehicle movement patterns within major china cities: Chengdu and Xi’an. For each city, we reserve 5,000 trajectories for testing. The introduction, statistics, and preprocess details of the datasets are summarized in Appendix~\ref{app:dataset}.

\noindent\textbf{Evaluation Metrics. }Following previous work\cite{du2023ldptrace}, We evaluate model performance by measuring Density Error, Trip Error, Length Error and Pattern Score. The details about these metrics are listed in Appendix~\ref{app:metrics}. For quantification, we partition each city into square grids of 50 meters, which corresponds to approximately $0.00045^\circ$ in longitude and latitude.

\noindent\textbf{Baseline Models.}
We compare our approach against several widely adopted convolution-based models in the domain of diffusion-based GPS trajectory generation. Specifically, we select three representative architectures:

\begin{itemize}[nosep, leftmargin=*]
\item \textbf{Traj-UNet} \cite{zhu2023difftraj}: A UNet-based model augmented with residual blocks and self-attention mechanisms to enhance feature extraction and sequence modeling capabilities.
\item \textbf{Geo-UNet} \cite{zhu2024controltraj}: An extension of Traj-UNet that introduces cross-attention layers in both the downsampling and upsampling paths to more effectively integrate conditioning information.
\item \textbf{WaveNet} \cite{kong2020diffwave, wei2024diff}: A convolutional model that employs stacked residual dilated convolution layers, enabling it to capture long-range dependencies and hierarchical temporal features.
\end{itemize}

Although Geo-UNet \cite{zhu2024controltraj} and Diff-RNTraj \cite{wei2024diff} incorporate road network information in their original implementations, we exclude this component in our experiments. Our goal is to evaluate the intrinsic capacity of each model to capture patterns from raw GPS trajectories alone. Accordingly, we adapt all baseline models to operate without access to road network data, ensuring a fair comparison under road network unaware settings.

\noindent\textbf{Training\footnote{Code is available here: \url{https://github.com/Zhiyang-Z/Traj-Transformer.git}}. }
The maximum diffusion timestep is set to 1000, and we sample every 5 steps in (\ref{eq:backward1}) during denoise. To ensure fair comparison, we apply identical training configurations across all models and do not perform any hyperparameter tuning. The training config is listed in Appendix~\ref{app:train_config} for reproducing.

\begin{table*}
   \centering
   \caption{Performance comparison of different model sizes \& embeddings (model symbols correspond to Table \ref{tab:model_configs}).}
   \renewcommand{\arraystretch}{1.2} 
   \begin{tabular}{l c c c c c c | c c c c } \hline
      \multirow{2}{*}{Methods} & \multirow{2}{*}{Size} & \multirow{2}{*}{Gflops} & \multicolumn{4}{c|}{Chengdu}& \multicolumn{4}{c}{Xi'an}\\ \cline{4-11}
                & & & Density ($\downarrow$) & Trip ($\downarrow$) & Length ($\downarrow$) & Pattern ($\uparrow$) &
                Density ($\downarrow$) & Trip ($\downarrow$) & Length ($\downarrow$) & Pattern ($\uparrow$) \\ \hline
         Traj-UNet \cite{zhu2023difftraj} & 16M & 1.6 & 0.0746 & 0.4488 & 0.0231 & 0.702 & 0.0682 & 0.4135 & 0.0277 & 0.714 \\
         Geo-UNet \footnotemark[3] \cite{zhu2024controltraj} & 8.2M & 0.6 & 0.0792 & 0.4528 & 0.0372 & 0.640 & 0.0739 & 0.5537 & 0.0364 & 0.690 \\
         WaveNet \footnotemark[3] \cite{kong2020diffwave, wei2024diff} & 5.1M & 0.4 & 0.0772 & 0.4436 & 0.0352 & 0.604 & 0.0752 & 0.5426 & 0.0331 & 0.704 \\ \hline
         T/lon-lat & 4.5M & 1.1 & 0.0568 & 0.3318 & 0.0189 & 0.760 & 0.0457 & 0.3221 & 0.0220 & 0.775 \\
         S/lon-lat & 8.5M & 2.1 & 0.0473 & 0.3040 & 0.0168 & 0.770 & 0.0395 & 0.2956 & 0.0219 & 0.788 \\
         B/lon-lat & 17M & 4.3 & 0.0376 & 0.2882 & 0.0143 & 0.812 & 0.0352 & 0.2431 & 0.0207 & 0.793 \\
         L/lon-lat & 33M & 8.5 & \textbf{0.0303} & \textbf{0.2688} & \textbf{0.0118} & \textbf{0.828} & \textbf{0.0295} & \textbf{0.1953} & \textbf{0.0189} & \textbf{0.830} \\ \hline
      \end{tabular}
   \label{Tab:results2}
\end{table*}

\begin{figure*}
   \includegraphics[scale=0.3]{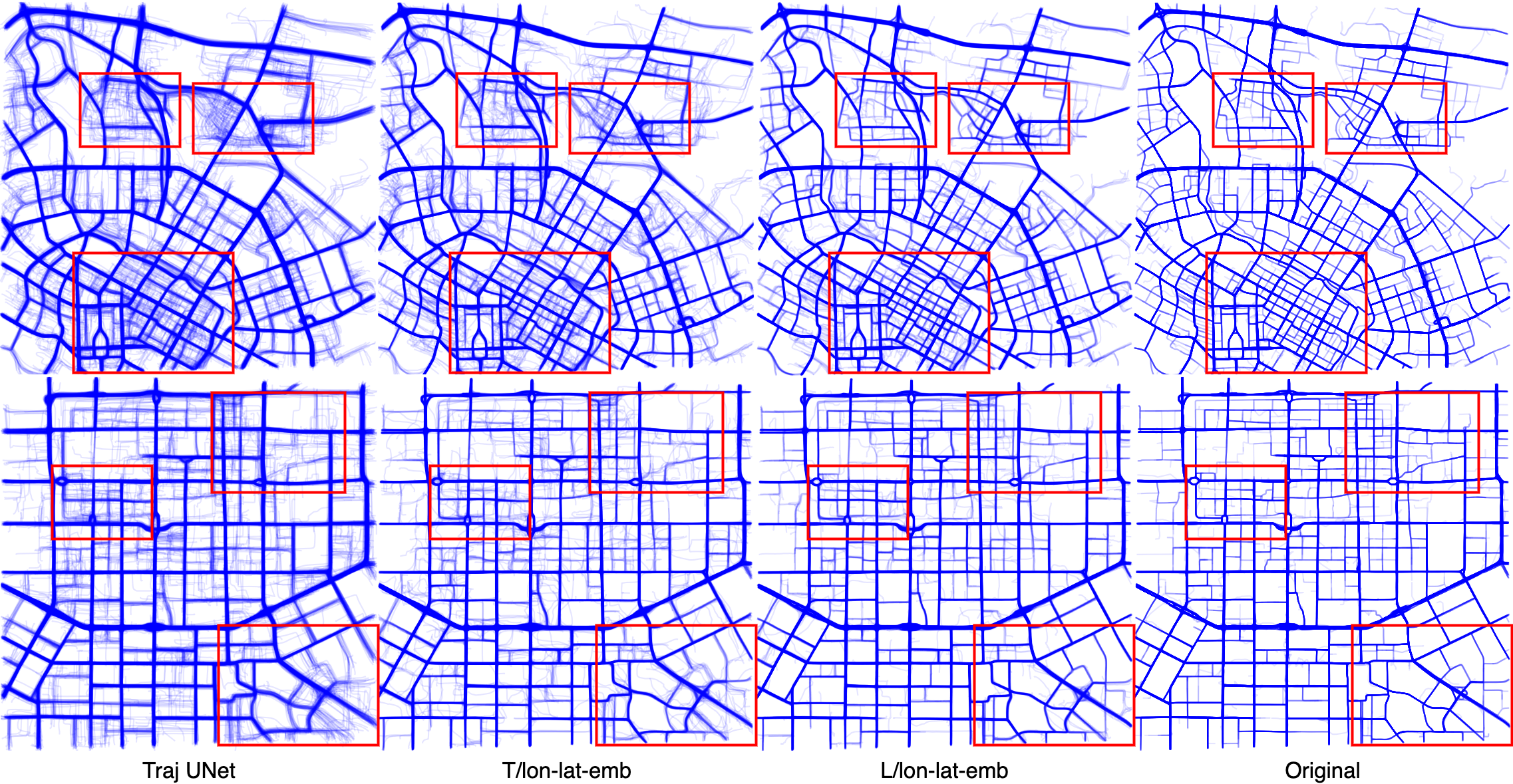}
   \centering
   \caption{Visualizations of two cities (top: Chengdu; bottom: Xi'an) from different models and original trajectory from test set (last column). Each subplot displays 5,000 trajectories generated from the test set without any visual post-processing. Red rectangles highlight high-density urban regions where differences in model performance are most evident. Lower-performing models struggle to preserve fine-grained street-level details in these areas, while higher-performing models more accurately capture and maintain the underlying street structures.}
   \label{compare2}
\end{figure*}

\section{Experiments}
We investigate the effects of different GPS embedding strategies and model sizes within our proposed model, and compare the results against convolution-based models. In this section, we refer to our models using a shorthand notation: model size followed by the embedding type. For example, \textbf{B/loc} denotes the Base-sized model using location embedding (loc-emb).

\subsection{GPS Embedding \& Model Scaling}
We conduct a comprehensive evaluation of model performance across various model sizes and embedding strategies. The full set of results is presented in Table \ref{Tab:results1}, where we report both model size and Gflops as measures of model complexity. As expected, larger models consistently yield better performance due to their higher representational capacity.

We further compare two embedding strategies—loc-emb and lon-lat-emb—within each model size category. Across all sizes, our experiments consistently show that lon-lat-emb leads to improved performance. This trend holds regardless of model size, underscoring the robustness and general effectiveness of the lon-lat-emb representation. To illustrate this, Figure \ref{embedding_training_curve} plots the density error as a function of training steps. At every point during training, models using lon-lat-emb achieve lower density errors than those using loc-emb, when model size is held constant. This consistent advantage highlights the superior capability of lon-lat-emb in capturing meaningful spatial features.

We attribute this performance gain to two main factors. First, lon-lat-emb preserves fine-grained spatial information by directly embedding raw geographic coordinates (longitude and latitude), enabling the model to better capture spatial distributions and patterns. Second, this representation allows attention mechanisms to operate jointly over spatial and temporal dimensions. In contrast, loc-emb typically restricts attention to the temporal dimension, limiting the model’s ability to capture complex spatial interactions.

In addition to evaluating numerical performance, we also examine the qualitative impact of model scaling and embedding choice. Figure \ref{compare1} presents visualizations of generated trajectories across models, ranging from the smallest configuration (T/loc-emb) to the largest (L/lon-lat-emb). The visual results clearly show that trajectory quality improves with both increased model size and richer spatial embeddings. In particular, larger models equipped with lon-lat-emb are significantly better at reducing trajectory deviations, especially in densely populated urban areas where spatial complexity is high. In these challenging regions, the smallest model (T/loc-emb) often fails to capture street-level details, leading to unrealistic or distorted trajectories. In contrast, the largest model (L/lon-lat-emb) produces more reliable and coherent paths, with a marked reduction in chaotic or implausible patterns. These findings highlight the importance of both model capacity and detailed spatial embeddings in enhancing trajectory generation fidelity.

Given that models using lon-lat-emb consistently outperform their loc-emb counterparts, we adopt lon-lat-emb as the default embedding configuration for subsequent model comparison.

\footnotetext[3]{Directly trained on raw GPS points. Road net embedding is elimnated.}

\begin{table*}
   \centering
   \caption{Density Error in Traffic Flow Distribution Prediction Across Time Periods (model symbols correspond to Table \ref{tab:model_configs}).}
   \renewcommand{\arraystretch}{1.2} 
   \begin{tabular}{l c c c c | c c c c } \hline
      \multirow{2}{*}{Methods} & \multicolumn{4}{c|}{Chengdu}& \multicolumn{4}{c}{Xi'an}\\ \cline{2-9}
                & 0AM-6AM & 6AM-12AM & 12PM-6PM & 6PM-12PM &
                0AM-6AM & 6AM-12AM & 12PM-6PM & 6PM-12PM \\ \hline
         Traj-UNet \cite{zhu2023difftraj} & 0.0980 & 0.1007 & 0.1216 & 0.1552 & 0.1089 & 0.1057 & 0.1189 & 0.1750 \\
         Geo-UNet \footnotemark[4] \cite{zhu2024controltraj} & 0.1026 & 0.1018 & 0.1208 & 0.1712 & 0.1131 & 0.1046 & 0.1146 & 0.1994 \\
         WaveNet \footnotemark[4] \cite{kong2020diffwave, wei2024diff} & 0.0982 & 0.1012 & 0.1213 & 0.1608 & 0.1128 & 0.1025 & 0.1155 & 0.1882 \\ \hline
         T/lon-lat & 0.0761 & 0.0769 & 0.0882 & 0.1272 & 0.0617 & 0.0565 & 0.0706 & 0.1231 \\
         S/lon-lat & 0.0579 & 0.0573 & 0.0736 & 0.1147 & 0.0570 & 0.0473 & 0.0572 & 0.1170 \\
         B/lon-lat & 0.0570 & 0.0570 & 0.0681 & 0.1044 & 0.0484 & 0.0451 & 0.0553 & 0.1116 \\
         L/lon-lat & \textbf{0.0560} & \textbf{0.0554} & \textbf{0.0680} & \textbf{0.1042} & \textbf{0.0481} & \textbf{0.0437} & \textbf{0.0553} & \textbf{0.1058} \\ \hline
      \end{tabular}
   \label{Tab:results2}
\end{table*}

\subsection{Model Comparison. } In this paper, our primary focus is to evaluate the impact of model capacity itself, independent of external information such as road network data. Consequently, when comparing with existing models—some of which were originally designed to incorporate external information—we do not strictly follow their original configurations. Instead, we retain their architectural designs while removing the modules responsible for integrating external data inputs. This approach enables a fair comparison that isolates the effect of model capacity free from the confounding influence of external data.

All quantitative results are summarized in Table \ref{Tab:results2}. Across all settings, our proposed models consistently outperform the convolution-based baselines, demonstrating clear advantages in both accuracy and spatial fidelity. Notably, our smallest model configuration, T/lon-lat, which uses only a quarter of the parameters and has comparable Gflops to the convolutional models, still achieves superior performance. This underscores the architectural efficiency of our model, delivering better results without increasing computational cost. The strong performance of T/lon-lat indicates that even at minimal scale, our model possesses a greater capacity to capture complex spatial patterns and accurately reconstruct fine-grained street-level structures.

To further validate these findings, we present qualitative visualizations of generated trajectories in Figure \ref{compare2}. These visualizations compare the outputs of our models with those of convolution-based models, alongside ground-truth trajectories from the test set. It is immediately clear that our models, even at small scales, recover detailed street geometries with remarkable accuracy. In contrast, the UNet-based models fail to preserve fine-grained structural details; their outputs often appear over-smoothed or geometrically inconsistent, especially in dense urban areas where complex road networks demand high spatial resolution for accurate representation.

This result, from quantitative metrics and qualitative visualization, strongly suggests that our approach is not only more parameter-efficient but also better at capturing spatial dependencies critical for accurate trajectory generation. Our architecture reconstructs complex trajectories more faithfully than UNet variants.

The advantages of our architecture become even more pronounced in the largest configuration, L/lon-lat. This model achieves the highest fidelity in trajectory generation, markedly reducing deviations from the ground truth and accurately recovering the underlying street network. By combining increased model capacity with the detailed lon-lat embedding, the model can effectively attend to both local and global spatial patterns, producing smoother and more realistic trajectories.

Overall, these results strongly validate the effectiveness of our model architecture and embedding strategy. They demonstrate that careful design choices in both model scaling and spatial encoding are crucial for high-quality trajectory generation—particularly in urban environments where preserving spatial detail is essential. Our findings also indicate that conventional architectures like UNet, even with ample computational resources, may fall short in tasks requiring fine-grained spatial reasoning.

\begin{figure*}
   \includegraphics[scale=0.26]{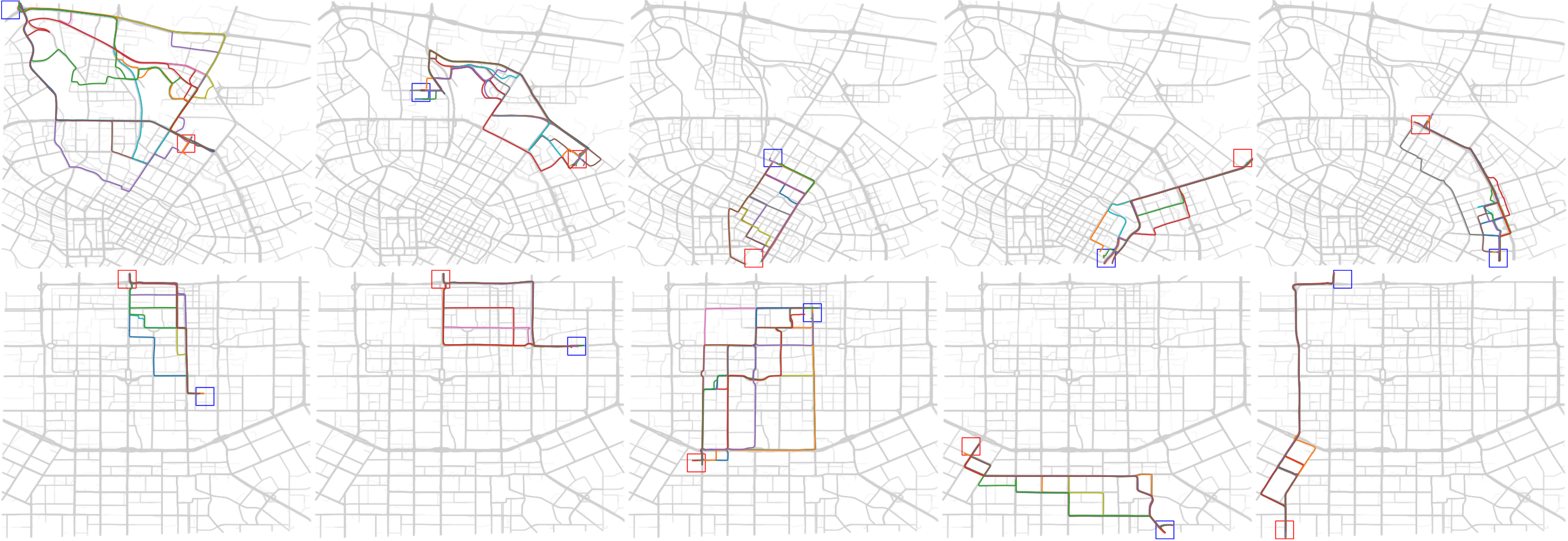}
   \centering
   \caption{Generated diverse trajectories in two cities (Top: Chengdu; Bottom: Xi'an). All trajectories share the same departure and destination regions. The red square marks the departure region, while the blue square indicates the destination region. The background shows the complete city map.}
   \label{diversity_xian}
\end{figure*}

\subsection{Utility of Generated Data \& diversity}
As the primary objective of trajectory generation is to facilitate the understanding and analysis of human mobility behaviors, evaluating the utility of the generated data is critical to determining the overall quality of the generative model. Beyond visual inspection or statistical similarity to real-world data, utility-focused evaluations provide insights into how well the generated trajectories can support real-world downstream applications. In this section, we conduct a utility assessment based on a representative downstream task—traffic flow prediction—which is a fundamental problem in intelligent transportation systems and urban computing.

To rigorously evaluate predictive performance, we first divide each day into four equal temporal segments, capturing morning, afternoon, evening, and night patterns, to account for diurnal variations in human activity. Simultaneously, the geographic area under study is partitioned in the same way as previous sections ($0.00045^\circ$ in both longitude and latitude), and then predict the total number of trajectories (i.e., traffic flow) occurring in each grid cell.

\footnotetext[4]{Directly trained on raw GPS points. Road net embedding is elimnated.}


As reported in Table \ref{Tab:results2}, our model consistently achieves the lowest density error across all time periods when compared to baseline models. This demonstrates the model’s enhanced capacity to reproduce realistic movement behaviors. These results underscore the practical utility of our approach, indicating that it is not only capable of generating real trajectories but also effective in supporting analytical tasks that depend on accurate mobility patterns.


Finally, we perform a controlled generation experiment in which specific conditions are fixed to guide the trajectory generation process. This setup allows us to evaluate the model’s capacity for diverse generation, thereby assessing whether it can produce varied trajectories that still conform to the same condition.
As demonstrated in our qualitative visualizations (\textbf{Figure} \ref{diversity_xian}), our model is capable of generating a wide range of distinct yet realistic trajectories under fixed conditions. This highlights its strength in learning a rich, multimodal distribution over human movement patterns, rather than collapsing into a single mode or deterministic output.

\section{Related Work}
\textbf{Diffusion Model:}
Generative models, e.g. GAN\cite{goodfellow2020generative}, VAE\cite{kingma2013auto, van2017neural},
normalizing flows\cite{rezende2015variational}, etc.,
have become widely used techniques for data generation.
Recently, diffusion models, a family of generative models, first proposed by Sohl-Dickstein et al.\cite{sohl2015deep} have
demonstrated remarkable performance in producing high-quality data.
Unlike GANs, diffusion models avoid adversarial training\cite{goodfellow2020generative} so that the training process becomes stable.
Diffusion models can operate in both discrete and continuous data spaces with either discrete or continuous timesteps,
leading to four possible categories.
DDPM\cite{ho2020denoising} perturbs data in continuous space by adding Gaussian noise to the original data $x_{0}$ and
the denoising process iteratively predicts $p(x_{t}|x_{t+1})$. This denoiseing process can also be derived by
distribution score estimation\cite{hyvarinen2005estimation,song2020sliced} and Langevin dynamics\cite{song2019generative}.
Furthermore, this perturbation can also be extended to continuous timestep and establish a connection with
Stochastic Diffierential Equation(SDE)\cite{song2020score}. In contrast, D3PM\cite{austin2021structured} applies perturbation in discrete data space using Markov process. 
Similar to continuous diffusion models, discrete diffusion can also be extended to continuous
timesteps\cite{campbell2022continuous, sun2022score} using the concept of discrete score estimation\cite{meng2022concrete, lou2023discrete}

\noindent\textbf{Trajectory Generation:}
Methods for trajectory generation can generally be categorized into two broad approaches: non-generative and generative.
Non-generative methods manipulate real trajectories by applying perturbations \cite{armstrong1999geographically, zandbergen2014ensuring} or combining segments from different real trajectories \cite{nergiz2008towards}. In contrast, generative methods employ neural networks to learn and sample from the underlying distribution of real-world trajectories. Generative Adversarial Networks (GANs) \cite{goodfellow2020generative} have been widely adopted for this task by representing trajectories as grid-based or image-like structures \cite{zhang2023dp, cao2021generating, ouyang2018non}. More recently, diffusion models have been introduced for trajectory generation by progressively perturbing data with Gaussian noise during training and reversing the process during sampling \cite{zhu2023difftraj}. ControlTraj \cite{zhu2024controltraj} improves upon DiffTraj \cite{zhu2023difftraj} by leveraging RoadMAE, a transformer-based autoencoder, to extract road segment embeddings and conditioning the generation process on this road-level information. Similarly, Wei et al. \cite{wei2024diff} use the Node2Vec algorithm to obtain road embeddings, which are then used in conjunction with a diffusion model to generate road-aware trajectories.


\section{Conclusions}
In this work, we propose Traj-Transformer, a transformer-based model for GPS trajectory generation. We explore two strategies for GPS point embedding, and experimental results show that separately embedding longitude and latitude yields better performance. Further evaluations demonstrate that our model effectively preserves fine-grained street-level details in dense urban areas—where convolutional approaches, such as UNet-based models, often struggle to capture the underlying structure. These results indicate that our model substantially improves generation quality and shows strong potential for generating realistic GPS trajectories without relying on auxiliary road information, thereby avoiding the complexity of multi-stage training pipelines \cite{zhu2024controltraj, wei2024diff}.

Beyond performance benefits, the transformer architecture provides a unified and flexible modeling framework. For example, as demonstrated in \cite{zhu2024controltraj}, embeddings from RoadMAE—a transformer-based autoencoder—can be seamlessly integrated into another transformer model to guide the reverse diffusion process. This demonstrates the feasibility of constructing an end-to-end, homogeneous pipeline using entirely transformer-based components.

\newpage



\bibliographystyle{ACM-Reference-Format}
\bibliography{sample-base}

\newpage
\appendix
\section{Datasets and Preprocess}
To comprehensively evaluate the performance of our proposed model, we conduct experiments on two large-scale GPS trajectory datasets\footnote{\url{https://outreach.didichuxing.com/}}, each capturing vehicle movement patterns within major china cities: Chengdu and Xi’an. These datasets contain extensive collections of recorded cab trajectories spanning two months—October and November of 2016. We merge the data from both months under the assumption that human mobility patterns remain relatively stable across a short time period (two consecutive months in the same year).

During preprocessing, we follow three key principles:
\begin{itemize}[nosep, leftmargin=*]
    \item Following \cite{zhu2023difftraj}, we remove all trajectories with lengths less than 120, as such short sequences may not provide sufficient information for effective learning.
    \item We remove trajectories containing consecutive GPS points with time gaps exceeding 25 seconds, which are indicative of GPS signal loss or logging interruptions.
    \item Following \cite{zhu2023difftraj}, we uniformly sample all remaining trajectories to a fixed length of 200 points to standardize the input sequence length.
\end{itemize}
For each city, we reserve 5,000 trajectories for testing. The statistical details of the datasets are summarized in Table~\ref{tab:datasets}, with all statistics computed after preprocessing.

\begin{table}
   \centering
   \caption{Statistics of Two Real-world Trajectory Datasets.}
   \renewcommand{\arraystretch}{1.2} 
   \begin{tabular}{ c c c c c } \hline
    City & Trajectory \# & Ave. Time & Ave. Distance\\ \hline
    Chengdu & 5779528 & 13.0 min & 4.8 km\\
    Xi'an & 3885527 & 14.6 min & 4.6 km\\ \hline
    \end{tabular}
    \label{tab:datasets}
\end{table}
\label{app:dataset}

\section{Evaluation Metrics}
We evaluate model performance by measuring similarity between generated trajectory distribution and real trajectory distribution.
Following previous work\cite{du2023ldptrace}, we use Jenson-Shannon divergence (JSD) to quantify:
$$JSD(P||Q) = \frac{1}{2}D_{KL}(P||M) + \frac{1}{2}D_{KL}(Q||M)$$
where $P, Q$ are real and generated trajectory distributions, $M = \frac{1}{2}(P + Q)$, and $D_{KL}$ represents KL divergence.

We partition each city into square grids of 50 meters, which corresponds to approximately $0.00045^\circ$ in longitude and latitude. For each grid cell, we compute the distribution of trajectory points located within it. Based on this spatial discretization, we calculate the following matrices:

\begin{itemize}[nosep, leftmargin=*]
   \item \textbf{Density error:} A global level metric that measures the similarity between entire generated and real trajectory.
   \item \textbf{Trip error:} A trajectory level metric that measures similarity of start/end points between generated and real trajectory.
   \item \textbf{Length error:} A trajectory level metric to evaluate the distribution of travel distances. It can be obtained by calculating the Euclidean distance between consecutive points.
   \item \textbf{Pattern score:} This is a semantic level metric defined as the top-n grids that occur most frequently in the trajectory.
   $$Pattern score = 2 \times \frac{Precision(P,P_{gen}) \times Recall(P,P_{gen})}{Precision(P,P_{gen}) + Recall(P,P_{gen})}$$
   where $P$ and $P_{gen}$ denote the original and generated pattern sets, respectively
\end{itemize}
\label{app:metrics}

\section{Training Config}
All experiments are conducted on NVIDIA A100 80GB. The maximum diffusion timestep is set to 1000, and we sample every 5 steps in (\ref{eq:backward1}) during denoise. We train all models using the AdamW optimizer with a constant learning rate of $1 \times 10^{-4}$, no weight decay and a batch size of 512. To ensure fair comparison, we apply identical training configurations across all models and do not perform any hyperparameter tuning. Each model is trained for approximately 500K-650K steps.
\label{app:train_config}

\end{document}